\documentclass[11pt,a4paper]{article}
\usepackage[hyperref]{acl2020}
\usepackage{times}
\usepackage{multirow}
\usepackage{latexsym}
\usepackage{makecell}
\usepackage{upquote}

\usepackage{booktabs}
\usepackage{graphicx}
\usepackage{microtype}
\usepackage{amsmath}

\title{A Survey on Graph Neural Networks for Knowledge Graph Completion}

\author{Siddhant Arora \\
  Indian Institute of Technology Delhi \\
  \texttt{siddhantarora1806@gmail.com} \\}

\date{}

\begin{document}
\maketitle
\begin{abstract}
Knowledge Graphs are increasingly becoming popular for a variety of downstream tasks like Question Answering and Information Retrieval. However, the Knowledge Graphs are often incomplete, thus leading to poor performance. As a result, there has been a lot of interest in the task of Knowledge Base Completion. More recently, Graph Neural Networks have been used to capture structural information inherently stored in these Knowledge Graphs and have been shown to achieve SOTA performance across a variety of datasets. In this survey, we understand the various strengths and weaknesses of the proposed methodology and try to find new exciting research problems in this area that require further investigation.
\end{abstract}

\section{Introduction}
Knowledge Bases are collection of factual information in the form of relational triplets. Each relational triplets can be represented as ($e_1$,r,$e_2$) where $e_1$ 
and $e_2$ are entities in knowledge base and r is the relation between $e_1$ and $e_2$. The most popular way of visualising knowledge base is by representing them as multi relational graph where each triplet ($e_1$,r,$e_2$) is represented as directed edge from $e_1$ to $e_2$ with label r. Knowledge Bases have used to improve performance across variety of tasks like Question Answering (\citet{KB-QA1}, \citet{KBQA2}), Dialogue Generation (\citet{KB-Dialogue}) and many others.

However, since Knowledge Bases are populated from automatic mining from texts, they are often incomplete since it is not possible to manually write all the facts, and there are often inaccuracies in extraction. This inaccuracy leads to a decline in performance across a variety of downstream tasks. Hence, there has been a lot of work in coming up with an efficient tool to complete the Knowledge Bases (KBs) by automatically adding new facts without requiring extra knowledge. This task is referred to as Knowledge Base Completion (or Link Prediction), where the goal is to solve queries like  ($e_1$,r,?).  

The first approach towards efficient Knowledge Base Completion were additive models like TransE (\citet{TransE}) and TransH (\citet{TransH}) where relations were interpreted as simple translations over hidden entity representations. Multiplicative models like Distmult (\citet{Distmult}) and Complex (\citet{Complex}) were then observed to outperform these simple additive models.  Instead of translation, RotatE (\citet{RotatE}) defines relation as simple rotations such that the head entity can be rotated in the complex embedding space to match the tail entity, which has been shown to satisfy a lot of useful semantic properties like compositionality of relations. Recently, more expressive Neural Network-based methods ( like ConvE (\citet{ConvE}) and ConvKB(\citet{ConvKB})) were introduced where the scoring function is learned along with the model. However, all these models process each triplet independently. As a result, these methods cannot capture semantically rich neighborhood and hence produce low-quality embeddings.

Graphs have been widely used to visualize real-world data. There has been tremendous progress in applying ML techniques over images and text, some of which are being successfully adapted to graphs (like \citet{GCN}, \citet{graphsage}, \citet{GAT}. Taking inspiration from this approach, a number of Graph Neural Network-based methods have been proposed to capture neighborhood in Knowledge Graphs for the KBC task. In this survey, we aim to look into some of these formulations.

\section{Dataset}

\begin{table*}[]
\caption{Summary of the datasets used in KBC experiments}
\label{tab:data}
\centering
\begin{tabular}{cccccc}
\toprule
Dataset          & Entities       & Relations & Train & Valid & Test \\ \midrule
FB15k-237          & 14,541 & 237 & 272,115 & 17,535 & 20,466          \\
FB15k         & 14,951 & 1,345 & 483,142 & 50,000 & 59,071          \\
WN18           & 40,943 & 18 & 141,442 & 5,000 & 5,000      \\
WN18RR           & 40,943 & 11 & 86,835 & 3,034 & 3,134     \\ \bottomrule
\end{tabular}
\end{table*}

The approaches for Knowledge Graph Completion have been evaluated on a variety of benchmarks datasets. Now, we will study in brief details about each of these large KGs used to assess performance in KBC tasks. Table \ref{tab:data} shows a description of the sizes of these benchmark datasets.

\begin{itemize}
    \item FB15k - FB15k is a subset of relational database Freebase and is a popular benchmark for Knowledge Graph Completion. In \citet{FB15k}, a serious issue was observed with this dataset. The dataset consist of inverse triplet like ($e_1,r^-1,e_2$) in test set such that ($e_2,r,e_1$) exists in the training set. Thus a simple model that can memorize the triplets in the training set can achieve very high test accuracy in this dataset. As a result, they proposed a new dataset Fb15k-237, where they had removed such inverse triplets.  
    \item WN18 - WN18 is a subset of WordNet KB containing lexical relation between words. Similar to FB15K, it contained inverse relations, and hence WN18RR was introduced. WN18RR has a hierarchical structure, and as a result, it poses a significant challenge to all KBC approaches that do not handle transitive relations.
\end{itemize}
Knowledge Graph Completion methods are evaluated using test queries ($e_1$,r,?), where we create a ranking list of all entities using the method's scoring function. We then compute metrics like Mean Reciprocal Rank that measures the average of the reciprocal rank of correct result $e_2$ for a given query, and Hits@N where measures number of times $e_2$ occurs in top N of the ranked list.

\section{Basic Concepts}
In this section, we will talk about some of the basic concepts related to Graph Neural Networks.

\subsection{Message Passing Neural Network}
Message Passing Network (\citet{Message_Passing}) is a framework that aims to generalize the various neural network proposed for graph-based data. Let us look at graph G with node embeddings $n_{v}$ for all nodes v, edge embeddings $e_{vw}$ for all edges between v and w. We first define message function as given below-
\begin{equation*}
    M(n_{v},n_{w},e_{vw})=f(n_{v},n_{v},e_{vw})
\end{equation*}
where f is nonlinear function and is usually taken as MLP. Then the net message sent to node v is computed by aggregating the message from all its neighbours w using the equation below-
\begin{equation*}
    {m^{t+1}}_{v}=AGG_{w \epsilon N(v)}(M({n^t}_{v},{n^t}_{w},{e^t}_{vw}))
\end{equation*}
where N(v) is neighbour set of v, t is timestep when messages are being aggregated and AGG is aggregation function that can be sum, mean or any other function. Finally, the embedding for node v are updated using the update function UPD as given below-
\begin{equation*}
    {n^{t+1}}_{v}=UPD({m^{t+1}}_{v},{n^{t}}_{v})
\end{equation*}

\subsection{Graph Convolution Network}
Graph Convolution Network is based on spectral graph convolution networks. Here the dataset consists of a graph with vertex set, adjacency matrix, and feature set for each node. These features could be categorical attributes like node labels, structural features like node degrees as well as the simple one-hot encoding for each node. GCNs aggregate information from neighbors of a node by using the simple equation given below.
\begin{equation*}
    {h_v}^k = \sigma (W_k (\Sigma_{u\epsilon N(v) \bigcup u} \frac{h_u^{k-1}}{\sqrt{\|N(u)\|\|N(v)\|}}))
\end{equation*}

\begin{figure}
    \centering
    \begin{minipage}{0.38\textwidth}
        \centering
        \includegraphics[width=\textwidth]{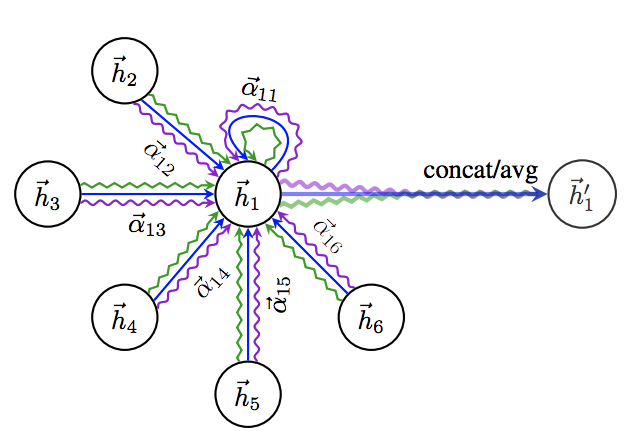}
    \end{minipage}
    \caption{Graph Attention Network methodology being shown over sample graph. Image from \citet{GAT}  }
    \label{fig:uniform}
\end{figure}

\subsection{Graph Attention Network}
In GCNs, all the neighbors contribute equally to the aggregation for each target node. Graph Attention Networks (GATs) [\citet{GAT}] overcome the shortcomings of the previous works by learning to assign varying levels of importance to all the nodes in the neighborhood of the node under consideration, rather than treating each node as equally important or using a fixed weight. Moreover, GATs are generalizable to unseen nodes (inductive learning), thus simulating a real-world setting and have been shown to advance state of the art across a variety of tasks.

\begin{table*}
\centering
\small

\begin{tabular}{ccccccc}
\hline
\makecell{\textbf{Model}} & \makecell{Uniform Aggregation\\ of Neighbours} & \makecell{Relation\\ Embedding} & \makecell{Use\\ Decoder} & \makecell{Incorporate\\ prior rules} & \makecell{Provide\\ Explanation} & \makecell{Use\\ Attribute} \\
\hline
\makecell{R-GCN \\ \citet{R-GCN}} & Yes & No & Yes & No & No & No \\
\makecell{TransGCN\\ \citet{Trans-GCN}} & Yes & Yes & No & No & No & No \\
\makecell{KB-GAT \\ \citet{KB-GAT}} & \makecell{No \\ (attention)} & Yes & Yes & No & No & No\\
\makecell{SACN\\ \citet{SACN}} & \makecell{No \\ (relation weighted)} & Yes & Yes & No & No & Yes\\
\makecell{ExpressGNN \\ \citet{GNNExpress}} & Yes & Yes & No & Yes & No & No\\
DPMPN \citet{DPMPN} & Yes & Yes & No & No & Yes & No \\
\hline
\end{tabular}
\caption{Summary of techniques used for Knowledge Graph Completion using GNNs using CNNs}
\label{tab:comparison}
\end{table*}
\normalsize
The input to GAT layer is a set of node features, $\pmb{x} = \{\vec{x_1}, \vec{x_2}, \cdots, \vec{x_N}\}$, where $N$ is the number of nodes. Then attention weight for each edge is calculated using the equation given below.

\begin{equation}
e_{ij} = a(\pmb{W}\vec{x_i}, \pmb{W}\vec{x_j})
\label{att}
\end{equation}

where W is weight matrix, a is any attention function, and $e_{ij}$ is attention weight for the edge between node i and node j. For each node i, the weights are then normalized by passing through softmax function. The output embedding for each node is then computed by the equation given below

\begin{equation}
\vec{x_i'} = \sigma(\sum_{j \in \mathcal{N}_i}\alpha_{ij}\pmb{W}\vec{x_j})
\label{update}
\end{equation}

\subsection{Decoder}
A lot of approaches that utilises GNN for KBC tasks uses them as auto encoder to help provide information of neighbouring relation triplets. Then, pre-existing scoring function like DistMult is enriched with this information in the form of initialisation of it's entity embeddings and outperforms the method that uses only scoring function. This scoring function is referred to as decoder in literature.
\section{GCN Encoder for Knowledge Graph Completion}
R-GCN (\citet{R-GCN}) was one of the earliest approaches to use GNNs for this task. They introduce a relational Graph Convolution Network, which produces locality-sensitive embeddings, which are then passed to the decoder that predicts missing links in KG. It is important to note that simple GCN cannot be used to embed KGs because it ignores the edge labels in the graph. As a result, the R-GCN slightly modified the scoring function of simple GCN to capture the relationship between the edges. 
\begin{equation*}
{h_v}^{k+1} = \sigma (\Sigma_{r\epsilon R} \Sigma_{j\epsilon {N_{v}}^{r}} \frac{1}{c_{r}}{W_r}^{k} {h_{j}}^{k} + {W_0}^{k} {h_{v}}^{k})  
\end{equation*}
Here R is a set of relations, and ${N_v}^r$ denote the set of entities that connected to v by relation r. We can see the model learns a new weight matrix for each relation, thus making their approach nonscalable for large graphs. They try to resolve this issue by coming with a basis and block diagonal decomposition. In the basis decomposition, each relation's weight matrix is represented as a sum of some base matrices. In the block diagonal decomposition, each relation weight matrix is represented as a direct sum over block-diagonal matrices. These approaches not only help to solve overfitting but also make relation weight matrices dependent on each other, thus helping to transfer knowledge learned from frequently occurring relation to learning weight matrices for rare relation better. They evaluate their approaches with state of the art multiplicative models like ComplEx and perform worse for FB15K and WN18 datasets. They then combine the R-GCN scores with DistMult scores and achieve comparable results to existing SOTA. The paper's main contribution is to show that GNNs can be successfully applied to KGs. However, they do not look into differential weighing of the node's neighborhood as well as require decoder since they do not learn relation embedding through GNN framework

\section{Relation Weighted GCN Encoder}
SACN (\citet{SACN}) try to extend on previous work by aggregating information from the node's neighborhood by being sensitive to edge relation types. This is referred to as Weighted Graph Convolution Network (WGCN). In WGCN, the whole graph is broken into subgraphs such that each subgraph contains edges of only one relation type. Then GCN is applied on each subgraph using the equation given below for relation t.

\begin{equation*}
    {h_v}^{k+1} = \sigma (\Sigma_{j\epsilon {N_v}} {\alpha_{t}}^{k} g({h_{v}}^{k},{h_{j}}^{k})) 
\end{equation*}

Here ${h_v}^{k+1}$ is embedding for node v at layer $k+1$, $N_v$ specifies the neighborhood of node v for relation type t and ${h_v}^{k}$ are node embedding at layer $k.$ g is simple matrix multiplication with connection coefficient matrix W which is shared across all relations which is in contrast to R-GCN which have separate weights for each relation. It then aggregates information across all these single relation subgraphs by weighing them with learnable parameter $\alpha_{t}$, which is different for each relation type t. Similar to the approach in R-GCN, SACN uses WGCN as an encoder to learn high-quality entity embeddings, which are then passed to decoder Conv-TransE. Conv-TransE is inspired by \citet{ConvE} 's success in applying convolution over embeddings. It, however, improves the pre-existing formulation by not reshaping relation and entity embedding vectors, thus helping to retain translation property of embeddings described in \citet{TransE}. Another interesting contribution of their work is that they also use node attributes to learn enhanced node embeddings by adding these node relation attributes as bridge nodes between the two entities connected by relation. The whole architecture is then learned end to end and is shown to achieve the state of the art results at that time over FB15k-237 and WN18RR datasets. Since most of the previous datasets do not consider the large number of entity attributes available, they come up with a new dataset, FB15k-237-Attr, that extracts the attribute triplets for entities in FB15k-237. They further show that their model can use this additional information to improve performance on link prediction further.
\section{Graph Attention Encoder}
Both of the two methods discussed above share the weakness of treating all neighboring nodes for each relation with equal importance. To overcome this limitation, KB-GAT (\citet{KB-GAT}) incorporates attention to identify important information in the neighborhood. Similar to previous methodologies, they learn GNN encoder followed by Neural Network-based scoring function as decoder i.e., ConvKB (\citet{ConvKB}). Unlike the earlier approaches, they use GNNs to learn both entity and relation embeddings. For each triplet ($e_i,r_k,e_j$), they represent a relation edge in the graph and compute representation for it using the equation given below.
\begin{equation*}
c_{ijk}=W_{1}[h_{i}\|\|h_{j}\|\|g_{k}] 
\end{equation*}
\begin{equation*}
 b_{ijk}=LeakyReLU(W_{2}c_{ijk})
\end{equation*}
Here, $h_{i}$, $h_{j}$ are entity embeddings and $g_{k}$ is relation embedding. These embeddings are first initialized by the additive TransE model. These embeddings are concatenated and multiplied by the learned weight matrix $W_{1}$ and $W_{2}$, before being passed through nonlinear activation. The weight matrix for all the edges to target node i are then passed through the softmax layer to produce attention weights for each edge.
\begin{equation*}
\alpha_{ijk}=\frac{exp(b_{ijk})}{\Sigma_{n\epsilon N_{i}} \Sigma_{r\epsilon R_{in}} exp(b_{inr}) } 
\end{equation*}
These $\alpha_{ijk}$ are used to weight the edge representations to compute output entity embeddings. Also, in order to not lose initial information, we add initial embeddings with the transformation to the output of the GAT network to produce final embeddings. Moreover, in each GAT layer, the output relation embedding is computed by matrix multiplication with a learned weighted matrix to input relation embedding. The paper also introduces the concept of auxiliary edges by adding direct edges to n-hop neighbors of the node where it experiments with 2 and 3 hop neighbors of a node. Finally, it learns the parameters of its GAT network by minimizing $h_{i}+g_{k}- h_{j}$, which is the scoring function introduced in TransE (\citet{TransE}). It then uses these embeddings to provide structural information to the decoder ConvKB and outperform SOTA approaches on FB15K-237, NELL-995, and Kinship dataset. Their approach, however, struggles to improve SOTA performance over WN18RR, which they believe can be accustomed to the fact that they do not handle hierarchical relations. They also performed ablation study over the hyperparameters and observed that auxiliary edges did not give much gain in performance. They also analyzed the distribution of attention weights and how this distribution varies with increasing iteration. Initially, as the epoch increases, the GAT layer gives more importance to direct neighbors and does not incorporate much information from distant neighbors. Eventually, in the later epochs, the model learns to capture the n-hop neighborhood of the node by assigning weights to auxiliary relations. They further analyzed how attention weights vary for different nodes by showing that for low in-degree nodes, effective embeddings can not be learned by simply aggregating information from direct neighbors, and hence higher attention weights are assigned by the model to n hop neighborhood. However, it is not clear why their approach is not trained end-to-end like earlier approaches. Moreover, since their GAT layer learns relation embedding, the motivation for using decoder is not very clear. It is also inferred in \citet{Evaluation}, that their experimental methodology is faulty, which will be discussed in more detail in later sections.

\section{Standalone GNN for KBC}
TransGCN (\citet{Trans-GCN}) is a GCN framework that jointly learns both entity and relation embeddings and hence dismisses the need for a decoder. It draws inspiration from R-GCN and aims to create a methodology that dismisses the need of task-specific decoder as well as avoids the computational cost of learning weights for relation twice- one during encoder step and second in the form of relation embedding in decoder step. It aims to use relation embedding to transform existing entity embedding for nodes in the graph such that it becomes a homogeneous graph. The transformation can be depicted more clearly by defining two transformation operators . and * as $t=h.r$ and $h=t*r$ as mentioned in \citet{Trans-GCN}. The GNN model's scoring function for each triplet is then defined by how close the transformed head entity is to the tail entity and vice versa. These transformation operators are used to define the node's neighbor by looking separately for incoming edges and outgoing edge. For an incoming edge (h,r,v), they transform entity h to $h.r$ and then include it in neighbourhood of v. Similarly for outgoing edge (v,r',t), they transform entity t to $t*r'$. Since edge labels are already taken into account by transforming entity embeddings, we can then apply simple GCN on the constructed homogeneous graph. Since there is no motivation for the output entity embeddings to still follow the scoring function with input relation embeddings, the input relation embedding is transformed by weight matrix to produce output relation embedding. The complete GNN framework is then optimized to minimize the scoring function using the last layer's entity and relation embeddings for valid triplets. They implement 2 GNN models- one which uses TransE (\citet{TransE}) scoring function called TransE-GCN and other which uses RotatE (\citet{RotatE}) scoring function called RotatE-GCN. Both TransE-GCN and RotatE-GCN outperform their base models TransE and RotatE correspondingly, and RotatE-GCN achieves SOTA performance across FB15k-237 and WN18RR. They further observed performance across nodes of varying degrees and showed that their performance struggled for node with low degrees where sufficient information is not available for learning useful embeddings and for nodes with a high degree where the model is not able to focus on important information for a particular node. It seems that both the above limitation can be solved using some of the methods introduced in KB-GAT. For nodes with low indegree, auxiliary edges can be added to incorporate more information. For nodes with high indegree, attention mechanisms can be used to learn important signals in the neighborhood.

\section{Incorporating rules with GNN}
\subsection{Markov Logic Networks}
\begin{figure}
    \centering
    \begin{minipage}{0.35\textwidth}
        \centering
        \includegraphics[width=\textwidth]{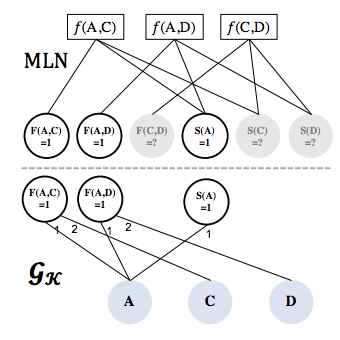}
    \end{minipage}
    \caption{MLN Network and KG being shown on top and bottom, respectively. Image from \citet{GNNExpress}}
    \label{fig:MLN}
\end{figure}

Markov Logic Networks (\citet{MLN}) acts as an interface between Applications like Robotics and Representation like Embeddings in AI. First Order Logic cannot be used as the interface since it is brittle and cannot handle uncertainty. Probabilistic Graphical Models that can handle uncertainty can't be used as an interface since they don't handle objects and relations. Markov Logic Networks act as the intersection of the above approaches and represent each Knowledge Base as (F,w) where F is set of formula and w is the weight assigned to each formula. Then a Markov Network is constructed from KG by making a bipartite graph with the formula on one side and their grounded instances on the other side, as shown in \ref{fig:MLN}. The formula and the corresponding ground predicates are connected by an edge, and then inference can be performed on this Markov Network. Then inference on Markov Network can be performed using the equation given below
\begin{equation*}
    {P_{w}}(O,H)=\frac{1}{Z(w)}exp(\Sigma_{f\epsilon F}w_{f}\Sigma_{a_{f}\epsilon A_{f}}\phi_{f}(a_{f}))
\end{equation*}
Here O are labelled facts, H are unlabelled facts, $A_{f}$ are grounded predicates, $\phi_{f}$ are assignment function that define truth value of grounded predicates and Z(w) is normalising factor
\subsection{MLN with GNN for inference}
As we can see in the equation in the previous section, computing Z(w) requires summing over an exponential number of ground predicates. In ExpressGNN (\citet{GNNExpress}, they use GNN to make this computation tractable. First of all, they represent KG as a bipartite graph with entities on one side and edges on the other side. They maximized the log-likelihood of observed facts ( referred to as "$P_w(O,H)$" in \citet{GNNExpress}) by trying to optimize the "variational evidence lower bound" of data. To do so, they computed posterior distribution, referred to as "$Q_{\theta}(H|O)$" in \citet{GNNExpress}, of hidden triplets using GNN to aggregate information from neighboring observed triplets. They optimize this objective using the Expectation Maximisation Algorithm. In the Expectation Step, the model learns the posterior distribution by trying to make it as close to likelihood distribution as possible since it is a good assumption to make that hidden fact follow the same distribution as observed ones. They also show that their approach can use the labeled data in the expectation step by including the maximization of posterior distribution of labeled facts in their optimization function. Finally, in the maximisation step, they used the new posterior distribution to update their likelihood estimates. Also, it is important to note that the posterior distributions are calculated with the help of GNN by equation given below-\\
\begin{equation*}
    Q_{\theta}((e_1,r,e_2))=\sigma(MLP(h_1,h_2,r))
\end{equation*}
where $(e_1,r,e_2)$ is a triplet, $h_1,h_2$ is embedding of the entities $e_1,e_2$, Also, it is important to know that entity embeddings were concatenation of GNN computed embedding formed by aggregation of messages from neighbours and tunable embedding which was learned exclusively using EM objective and provided the model with additional flexibility. Finally, they test their approach on KBC tasks by studying link prediction performance on FB15k-237 for 2 model variants, namely ExpressGNN-E, which used only Expectation step and ExpressGNN-EM, which used the complete EM algorithm. They observed SOTA performance outperforming the existing models by a huge margin. They also observed that their model performed very well even with low amounts of training data since they can incorporate logical rules in the formulation. Finally, they also showed that their model is more capable of performing on relations with very low observed data since the model is able to use logical knowledge to make informed predictions. 
\section{Explainable predictions with GNN}
DPMPN (\citet{DPMPN}) aims to construct query dependent subgraph that provides some explanation about the prediction made by the GNN model, thus helping us to reason about predictions rather than treating the GNN model as a black box. Moreover, in these subgraphs, we could use attention weights on each node to perform differential coloring of each node, which can show which nodes were important for prediction and provide intuition on how information travels from head entity to tail entity. They take inspiration from \citet{Bengio_cons}, where their modeling consists of 2 GNNs - the first GNN operates on the whole graph and tries to learn query independent properties of the graph. In contrast, the second GNN operates on the subgraph surrounding the head entity and provides query specific information as well as construct subgraph that provides reasoning for predictions. The first global GNN is referred to as "Inattentive GNN" (IGNN), whereas the second GNN that looks at query local features is referred to as "Attentive GNN" (AGNN) in \citet{DPMPN}. 
\subsection{IGNN}
To make IGNN scalable, there is only one forward pass through IGNN in batch since it is query invariant. The message passing algorithm (\citet{Message_Passing}) is repeated for t timesteps where the hidden state is updated at each time step, and then the hidden state at the last time step is used for downstream models i.e., AGNN and attention module. 

\subsection{AGNN}
In AGNN, they run GNN on subgraph constructed based on input query. For the query  (e1,r,?), they start from subgraph that contains only the entity e1. They then expand this subgraph by adding neighbors of entity e1 and so on. They use three sampling strategies. First, to construct a subgraph at timestep t+1, they sample a subset of nodes from subgraph at time t, which is known as attending-from horizon denoted by $\delta\widehat{G_{t}}$. They use attention scores from timestep t to sample top k nodes with the highest attention weight. We will discuss how they compute these attention scores ($a^t$) using the attention module in the next subsection. They then look at neighbors of these sampled nodes which are not already included in subgraph at timestamp t. They then sample nodes from this neighborhood, which is referred to as the sampling horizon. Now they want to further sample from this sampling horizon, for which they use attention scores ($a^{t+1}$) to samples top K nodes from sampling horizon to produce attending to-horizon. They then construct edges to these sampled nodes to aggregate messages for GNN at timestep t. Further, they use the attention scores to update the hidden states for node and then implement the Message Passing network inspired from \citet{Message_Passing}.
\subsection{Attention module}
 The attention module first computes the transition matrix T, which captures the interaction between every pair of nodes. For the pair (v,v'), they compute the net interaction as a sum of 2 interactions- the first interaction is between the underlying representation of v and v' for query dependent AGNN network, which measures similarity between 2 nodes close to query. The other is between the query-specific AGNN representation of v and IGNN representation of v' to capture the similarity of nodes that were never included in the subgraph but could play a role in predicting the tail entity. They then compute output attention weights from the equation given below-
\begin{equation*}
    a^{t+1}=\frac{{T^t}{a^t}}{\|{T^t}{a^t}\|}
\end{equation*}
 At the last time step, these attention scores are used as inference probabilities for the query.
  \begin{table*}[t!]
\centering
\small
{\begin{tabular}{lcccccccc}
\toprule
\textbf{Method} & \multicolumn{2}{c}{\textbf{FB15k-237}} & \multicolumn{2}{c}{\textbf{WN18RR}} & \multicolumn{2}{c}{\textbf{FB15k}} & \multicolumn{2}{c}{\textbf{WN18}}\\
\cmidrule(lr){2-3} \cmidrule(lr){4-5} \cmidrule(lr){6-7} \cmidrule(lr){8-9}
 & MRR & H@10 & MRR & H@10 & MRR & H@10 & MRR & H@10  \\ 
\midrule
ComplEx & 0.25 & 0.43 & 0.44 & 0.51 & 0.692 & 0.840 & 0.941 & 0.947 \\
R-GCN & - & - & - & - & 0.651 & 0.825 & 0.814 & 0.955 \\
R-GCN+ & - & - & - & - &  0.696 &  0.842 & 0.819 & 0.964 \\
TransE-GCN & 0.315 & 0.477 & 0.233 & 0.508 & - & - & - & -\\
RotatE-GCN & 0.356 & 0.555 & 0.485 & 0.578 & - & - & - & -\\
KB-GAT* & 0.518 & 0.626 & 0.440 &  0.581 & - & - & - & -  \\ 
SACN & 0.35 & 0.54 & 0.47 & 0.54 & - & - & - & -  \\
SACN-Attr & 0.36 & 0.55 & - & - & - & - & - & -  \\
DPMPN & 0.369 & 0.53 & 0.482 & 0.558 & - & - & - & -\\
ExpressGNN & 0.49 & 0.608 & - & - & - & - & - & - \\ 

\bottomrule
\end{tabular}}
\caption{Overall performance of all models in KG completion task using the Hits@10 and MRR reported in corresponding papers. * There is error in evaluation protocol of KB-GAT as well as test data leakage as shown in \citet{Evaluation}.}
\label{f1_scores}
\end{table*}
 
 \subsection{Analysis}
The model attains slightly worse results to SOTA performance on MRR for WN18RR and FB15k-237. However, they achieve very good performance for Hits@1 and Hits@3, showing that their model is very good at exact predictions. Moreover, they perform ablation study where they show that sampling more node for attending from horizon always give some gain in performance. However, time cost also increases tremendously in sampling more nodes for attending from the horizon. As a result, they have to restrict themselves to sampling only a few nodes to make their approach scalable.
 
\section{Result}
Complex results for FB15k-237 and WN-18RR are taken from \citet{SACN},\citet{DPMPN} and for FB15k and WN18RR are from (\citet{R-GCN}). R-GCN (\citet{R-GCN}) results are comparable to complex when used in conjunction with DistMult scores (R-GCN+), thus it is inferior in performance to other GNN based models which outperform ComplEx. ExpressGNN is the best performing in Fb15k-237, which shows huge gain can be made in performance by incorporation logic rules in the embedding based framework. Moreover, it can also be seen that incorporating the abundant entity attribute information can give some increase in performance. Also, for WN18RR, RotatE-GCN seems to be the best performing model since WN18RR has a lot of symmetric relations. All models other than Rotate-GCN cannot capture symmetric relations very well; however, as shown in \citet{RotatE}, Rotation transformations can infer the symmetry/antisymmetry pattern. 
\subsection{Incorrect Evaluation Methodology}
In \citet{Evaluation}, they observed that for some KBC methods like CapsE, ConvKB, and KB-GAT, the resulting score distribution was very different from those obtained by earlier proposed scoring function like ConvE. They found out that a lot of triplets have the exact same score. They traced this unusual behavior to lot neurons becoming inactive due to RELU activation function used in these ranking methods. Thus, to reason about the performance of these methods, it became imperative to understand how these methods deal with similar scores. They observed that KB-GAT's reported scores are observed by placing the correct triplet at the beginning. If the correct triplet is placed randomly in the ranked list, then we find a sharp decline in performance. As a result, we have excluded KB-GAT from our discussions and are interested in looking into evaluation methodology followed for other ranking functions.

\section{Concluding Remarks}
We can observe Graph Neural Networks have been widely adopted to improve SOTA performances for KBC tasks. However, this success has not been limited to only link prediction tasks, but there is a lot of active research is using GNNs on Knowledge Graphs for other related tasks. Recently,\citet{Heuristics_LP} has used GNNs to predict heuristics that can better help estimate the likelihood of 2 nodes to be connected to each other. Moreover, all Knowledge Graph Completion methods make an inherent assumption that all the entities in test triplets have been seen in training time, but this assumption might not hold in real-world scenarios. \citet{KBC_OOV} look into solving this problem by using GNN to compute a representation of entities, not seen in training time, from their neighborhood at test time. \citet{Logic_KBC_OOV} extends this idea by using rules along with GNN for the above task. \citet{OAG} utilizes GNN to perform entity linking over large KGs. 

As can be seen in Table \ref{tab:comparison}, there is no single methodology that is looking into all the benefits that can be achieved by GNN. ExpressGNN and DPMPN networks don't seem to be completely handling heterogeneity in the graph. When they aggregate messages from their neighborhood, although the relation embeddings are an input to compute messages from each neighbor, they do not weigh these messages based on the relation like in KB-GAT or SACN. Also, it would be interesting to see if some of the DPMPN ideas of explainability can be applied to ExpressGNN easily. Finally, it has been argued that a lot of models struggle to perform over hierarchical graphs like WN18RR. This is because none of the GNN based approaches so far have looked into effectively handling hierarchical relations. \citet{MuRP} has tried to project KGs into hyperbolic space for this purpose, but it still looks at each triplet independently. Future work on using GNN on hyperbolic space should improve state of the art over hierarchical graphs like WN18RR.  

\bibliography{anthology,acl2020}
\bibliographystyle{acl_natbib}

\end{document}